\documentclass{article} 
\usepackage{iclr2026_conference,times}

\usepackage{amsmath,amsfonts,bm}

\def\eqref#1{equation~\ref{#1}}

\def\1{\bm{1}}

\DeclareMathAlphabet{\mathsfit}{\encodingdefault}{\sfdefault}{m}{sl}
\SetMathAlphabet{\mathsfit}{bold}{\encodingdefault}{\sfdefault}{bx}{n}

\usepackage{hyperref}
\usepackage{url}
\usepackage{tcolorbox}
\tcbuselibrary{breakable}
\usepackage{etoolbox}
\usepackage{booktabs}
\usepackage{multirow}
\usepackage{makecell}
\usepackage{graphicx}
\usepackage{enumitem}
\usepackage{cleveref} 
\usepackage{subcaption}
\usepackage{amsmath}

\usepackage[T1]{fontenc}
\usepackage[utf8]{inputenc}
\usepackage{wrapfig}

\title{TPS-Bench: Evaluating AI Agents’ Tool Planning \& Scheduling Abilities in Compounding Tasks}

\author{
Hanwen Xu, Xuyao Huang, Yuzhe Liu, Kai Yu, Zhijie Deng 
\thanks{Corresponding author.} \\
Shanghai Jiaotong University\\ 
\texttt{\{dctnorin, huangxuyao, 2263692082, kai.yu, zhijied\}@sjtu.edu.cn} \\ 
}

\newtcolorbox{promptbox}[1][]{
  colback=gray!10,    
  colframe=black!50,  
  title=#1,           
  fonttitle=\bfseries\sffamily,   
  fontupper=\sffamily,            
  fonttitle=\bfseries,
  boxrule=0.5pt,
  arc=2pt,
  left=5pt,right=5pt,top=5pt,bottom=5pt
}

\iclrfinalcopy 

\makeatletter
\def\ps@headings{%
  \let\@oddhead\@empty
  \let\@evenhead\@empty
  
}
\makeatother

\begin{document}

\maketitle

\begin{abstract}
Large language model (LLM) agents have exhibited strong problem-solving competence across domains like research and coding.
Yet, it remains underexplored whether LLM agents can tackle compounding real-world problems that require a diverse set of tools to complete. 
Given a broad, heterogeneous tool repository, LLM agents must not only select appropriate tools based on task planning analysis but also strategically schedule the execution order to ensure efficiency.
This paper introduces \textbf{TPS-Bench} to benchmark the ability of LLM agents in solving such problems that demand \textbf{T}ool \textbf{P}lanning and \textbf{S}cheduling. 
TPS-Bench collects 200 compounding tasks of two difficulty levels, based on a tool repository containing hundreds of model context protocol (MCP) tools. 
In particular, each task is composed of multiple subtasks, such as web search, map navigation, calendar checking, etc., and each subtask can be completed by a basic tool. 
Our evaluation emphasizes both task completion rate and efficiency. 
The empirical studies on popular closed-source and open-source LLMs indicate that most models can perform reasonable tool planning, but differ in scheduling. 
For example, GLM-4.5 achieves an outperforming task completion rate of 64.72\%  with extensive sequential tool calls, hence suffering from significantly long execution time. 
By contrast, GPT-4o prioritizes parallel tool calls but achieves only a 45.08\% completion rate. 
Considering reinforcement learning (RL) can be a viable way to improve the scheduling efficiency without compromising performance, we perform an initial study on Qwen3-1.7B and witness a 14\% reduction in execution time alongside a 6\% gain in task completion rate based on rarely 100 RL training samples. Our code is available \href{https://github.com/hanwenxu1/mcp-agent}{https://github.com/hanwenxu1/mcp-agent}.
\end{abstract}

\section{Introduction}

Large language model (LLM) agents have demonstrated strong capabilities in solving various problems that require tool use~\citep{NEURIPS2024_2db8ce96, qin2023toolllmfacilitatinglargelanguage, nakano2022webgptbrowserassistedquestionansweringhuman}
, including deep research~\citep{openai2025-deep-research}, coding~\citep{yang2024sweagentagentcomputerinterfacesenable}, and data analysis~\citep{hong2024datainterpreterllmagent}, etc.
For example, GPT-o3~\citep{openai_o3_system_card_2025} can suitably solve the web search tasks in AssistantBench~\citep{yoran2024assistantbenchwebagentssolve} using the Google search tool, and GPT-5~\citep{openai-gpt5} excels in using edit tools to tackle coding tasks in SWE-Bench~\citep{jimenez2024swebenchlanguagemodelsresolve}. 

\begin{figure}[t]
    \begin{center}
        \includegraphics[width=1.0\textwidth]{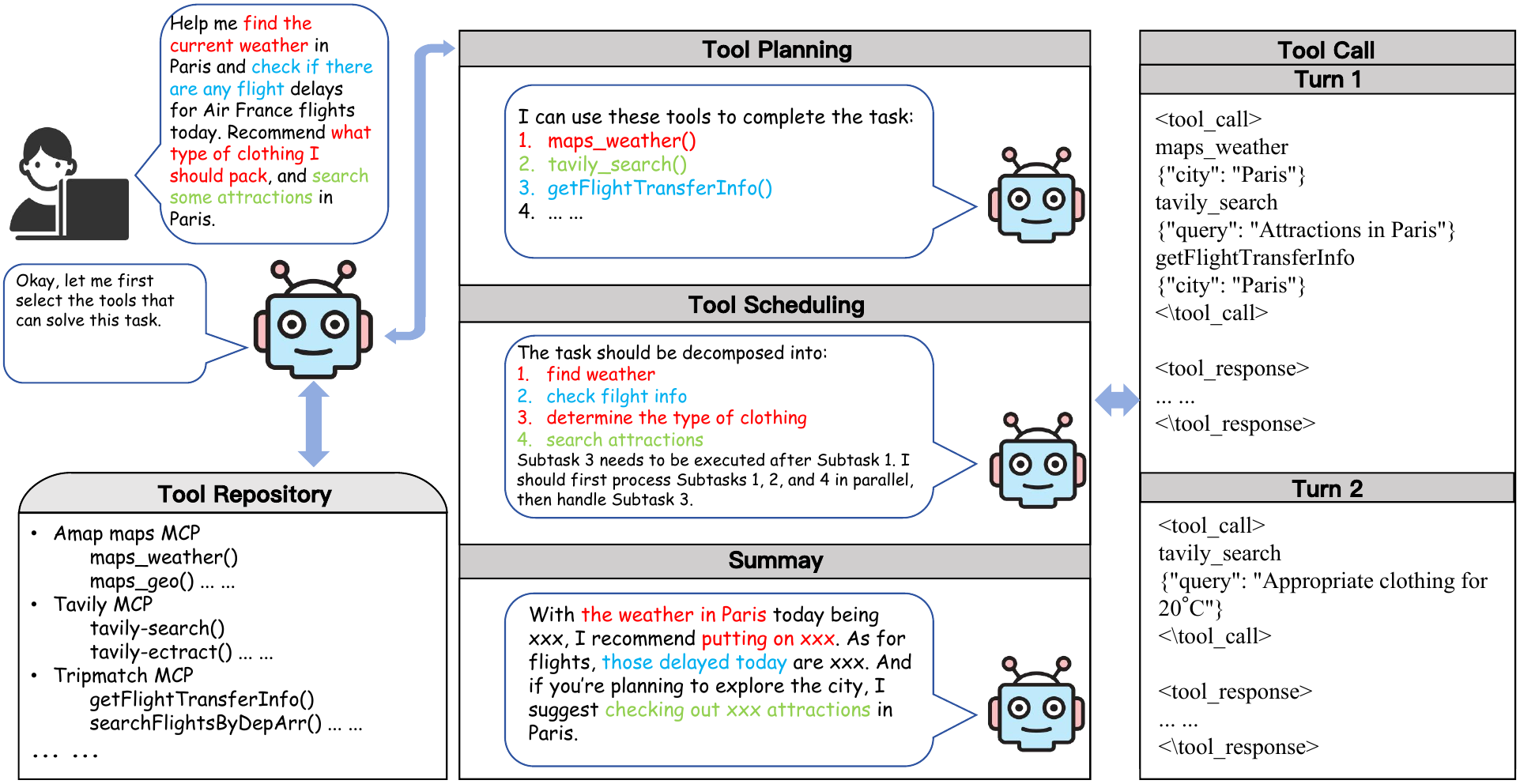}
    \end{center}
    \vspace{2em}
    \caption{TPS-Bench assesses the tool planning and scheduling ability of LLM agents for solving compounding tasks. As shown, the LLM agent needs to first select tools capable of solving the task from a tool repository. Next, the LLM agent decomposes the original task into multiple subtasks and identifies their dependency relationships (subtasks that have no dependencies are marked by different colors in the figure). After that, the LLM agent performs tool calls and collects the tool responses in multiple turns, after which the final answer is delivered. 
    }
    \label{fig: TPS-Bench}
\end{figure}

However, real-world scenarios often involve \emph{compounding tasks} that demand the collaboration of various tools, which present new challenges to LLM agents. 
As exemplified in Figure~\ref{fig: TPS-Bench}, a compounding task can involve multiple subtasks, each of which can be addressed by a basic tool.
For example, the request ``\emph{Please check the dates of upcoming holidays, recommend travel destinations for me, list a detailed itinerary, and arrange hotel reservation.}'' can be decomposed into subtasks such as calendar checking, web search, itinerary planning, hotel reservation, etc.
Tackling such a task requires LLM agents not only to engage in task comprehension and tool planning to identify necessary tools for completing the task, but also to employ strategic scheduling so that independent subtasks can be parallelized while dependent ones adhere to a sequential workflow. 



This paper introduces \textbf{TPS-Bench} for assessing the \textbf{T}ool \textbf{P}lanning and \textbf{S}cheduling ability of LLM agents. 
It consists of 200 compounding tasks defined on hundreds of model context protocol (MCP) tools.
The involved subtasks include weather report, calendar checking, web search, map navigation, etc.
According to compounding complexity, 
we organize TPS-Bench into two levels: 
TPS-Bench-Easy, which features simple and weakly relevant subtasks such as ``\emph{check tomorrow's weather and recommend autumn-appropriate recipes...}'', and
TPS-Bench-Hard, which comprises subtasks more compactly and implicitly: ``\emph{check the price fluctuations of Apple phones over the past year and recommend phones under 1,000 US dollars...}''.
Considering the involved tasks are open and no standard answers exist,
we use LLM-as-a-judge~\citep{zheng2023judgingllmasajudgemtbenchchatbot} to evaluate task completion rate. 
In addition, we systematically measure token usage and time consumption to assess the efficiency. 

We perform extensive empirical studies with closed-source and open-source LLMs, including GPT-4o~\citep{openai2024gpt4o}, Kimi-K2~\citep{kimiteam2025kimik2openagentic}, DeepSeek-R1~\citep{deepseekai2025deepseekr1incentivizingreasoningcapability}, GLM-4.5~\citep{zeng2025glm45}, QwQ-32B~\citep{qwq32b}, Qwen3-32B~\citep{qwen3}, and Qwen3-1.7B~\citep{qwen3}. 
We notice that most models exhibit reasonable performance in tool planning but behave differently in scheduling. 
E.g., GLM-4.5 tends to employ sequential tool execution, achieving the highest task completion rate of 64.72\% for TPS-Bench-Hard, at an average cost of 217.8s and 14k tokens per query.
In contrast, GPT-4o prioritizes parallel scheduling, which leads to 76.84s and 9k tokens per query, 
but the task completion rate is rarely 45.08\%.
Besides, Qwen3-32B stands as a strong open-source baseline for TPS-Bench. 
It consumes 8.0k average token usage and achieves a 56.72\% task completion rate on TPS-Bench-Hard.


We then investigate reinforcement learning (RL)-based post-training to improve the task scheduling ability of LLMs for both task completion and efficiency, considering its 
effectiveness displayed in agentic workflows~\citep{zhang2025aflowautomatingagenticworkflow}. 
In particular, we construct a training set of 100 samples, denoted as TPS-100, and use Group Relative Policy Optimization (GRPO)~\citep{shao2024deepseekmath} to train the cost-effective Qwen3-1.7B~\citep{qwen3}.
With only 125 training iterations, we witness a 6\% improvement in task completion rate alongside a 14\% reduction in execution time. 
We will open-source both the benchmark and training dataset once acceptance.

\section{Related Work}

\paragraph{Evaluation of LLM-based agents.} 
As LLMs pave a promising path for the development of general AI agents~\citep{xi2025rise}, recent years have witnessed a surge in benchmarks designed to evaluate the capabilities of LLM-based agents. These efforts span multiple dimensions: some studies assess agents' general problem-solving abilities across diverse environments~\citep{liu2023agentbench, yu2025c, li2024embodied, trivedi2024appworld, barres2025tau2benchevaluatingconversationalagents} for broad-capability assessment. Some focus on evaluating LLM tool use effectiveness~\citep{huang2023metatool, chen2023t, ye2024tooleyes} as a key interactive capability. Others specialize in researching domain-specific agents' abilities~\citep{jimenez2023swe, guo2024ctooleval, xie2024travelplanner} for professional-grade performance. However, these benchmarks primarily emphasize task completion effectiveness but pay limited attention to the efficiency metrics, such as tool usage, token consumption, and execution time, which are significant factors for the cost-effectiveness and scalability of agents in real-world tasks.

\paragraph{Efficiency in AI Agents.}
Efficient Agents~\citep{wang2025efficientagentsbuildingeffective} is closest to our work. However, it concentrates exclusively on developing strategies to enhance the operational efficiency of agents in single-task or limited-tool scenarios. This gap is particularly critical as currently agents increasingly operate in strictly token-constrained or latency-sensitive environments~\citep{miao2023towards}. Recent studies have introduced various methods for resource-aware inference and efficiency optimization, such as adaptive computation~\citep{zhou2024survey}, cost-efficient tool use, and latency-aware decoding~\citep{wang2025faster}. Our work further fills the aforementioned benchmark gap by proposing a dedicated efficiency-oriented benchmark system, which is specifically tailored to the unique functional demands and operational characteristics of our target multi-task, multi-tool agent settings.

\paragraph{Benchmark Construction and Task Designation.}
Several studies have explored benchmarks and task design for evaluating LLM agents. BIG-bench~\citep{srivastava2023beyond} provides diverse and challenging tasks for multi-faceted evaluation. SmartPlay~\citep{wu2023smartplay} takes a game-based approach to evaluate LLM agents across 6 distinct games, each designed to challenge various key capabilities. More recently, AgentBoard~\citep{chang2024agentboard} introduced a unified benchmark for multi-turn LLM agents, featuring 9 diverse tasks and various environments that require partially observable and multi-round interactions. However, these benchmarks predominantly rely on tasks with isolated objectives or simple sequential structures, thereby failing to capture real-world tasks' potential interdependencies, parallelization opportunities, and the compounding nature of problems  that necessitate sophisticated tool planning and scheduling strategies. To address this gap, TPS-Bench explicitly incorporates compound tasks with intricate subtask dependencies and clear parallelization potential, requiring agents not only to select appropriate tools but also to strategically schedule their execution for optimal efficiency. Furthermore, TPS-Bench introduces graded task difficulty levels—a feature notably absent in existing benchmarks.

\paragraph{MCP for Tool-Augmented Agents.}
The model context protocol (MCP), a framework designed to bridge the gap between LLM agents and external systems, has emerged as a pivotal standard for enabling seamless and dynamic interactions between LLM agents and external tools, thereby expanding the scope and complexity of tasks agents can undertake~\citep{singh2025survey, ray2025survey}. By facilitating this tight integration, MCP has expanded the scope and complexity of tasks. Although potential security concerns remain, recent studies have outlined future directions to address these challenges, paving the way for more robust and trustworthy MCP deployments~\citep{hou2025model}. In line with these key developments in the field, our work applies diverse MCP tools across categories, constructing a realistic, heterogeneous tool-fetching environment that closely mirrors complex real-world application scenarios. By strategically integrating MCP tools with a wide range of distinct, complementary functionalities, TPS-Bench provides a solid framework for assessing and advancing tool-augmented agents while reinforcing tool-planning and scheduling capabilities.

\begin{figure}[t]
  \centering
  \begin{subfigure}{0.45\textwidth}
    \centering
    \includegraphics[width=\linewidth]{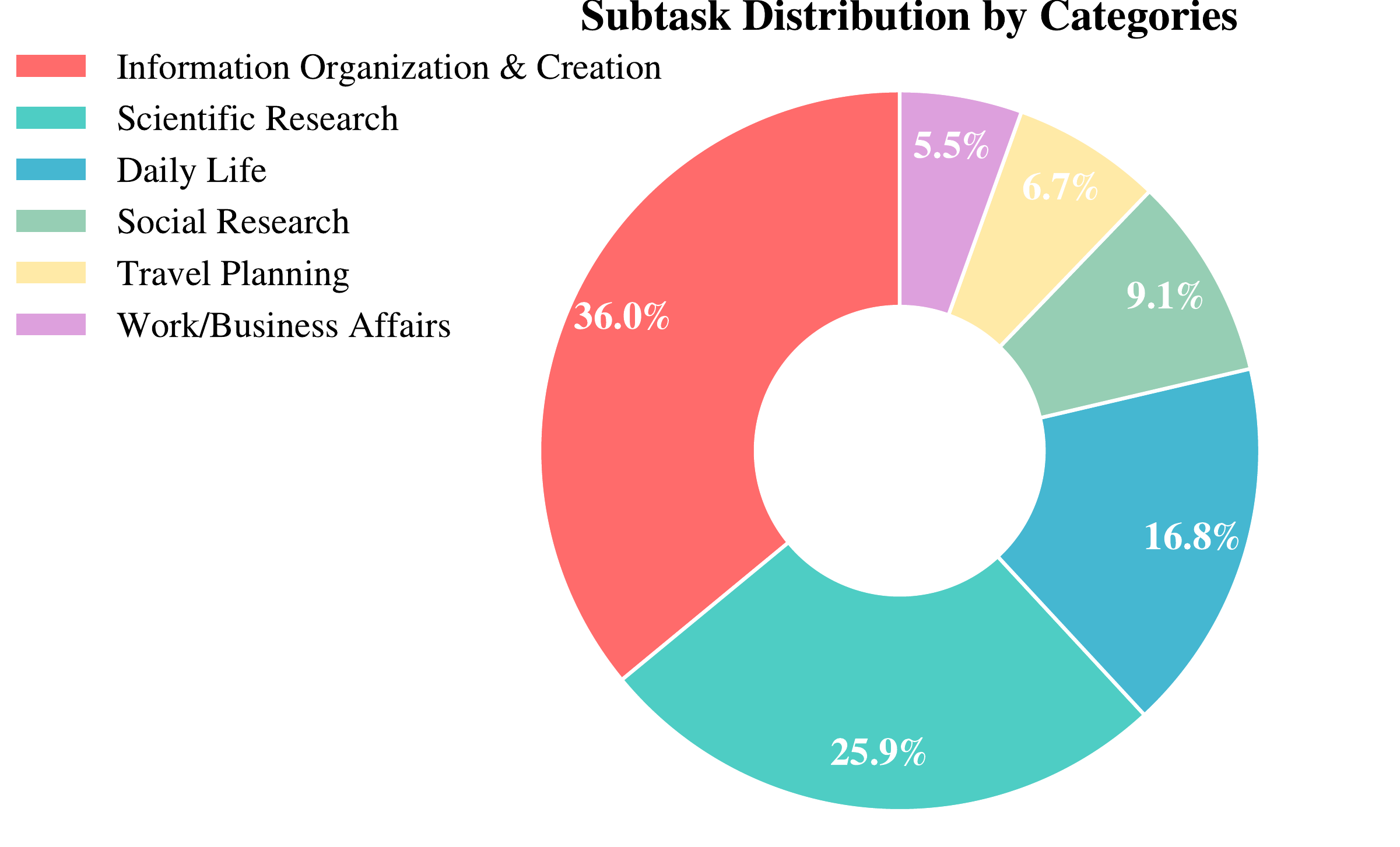}
    \label{fig: subtask+tool: subtask}
  \end{subfigure}
  \hfill
  \begin{subfigure}{0.45\textwidth}
    \centering
    \includegraphics[width=\linewidth]{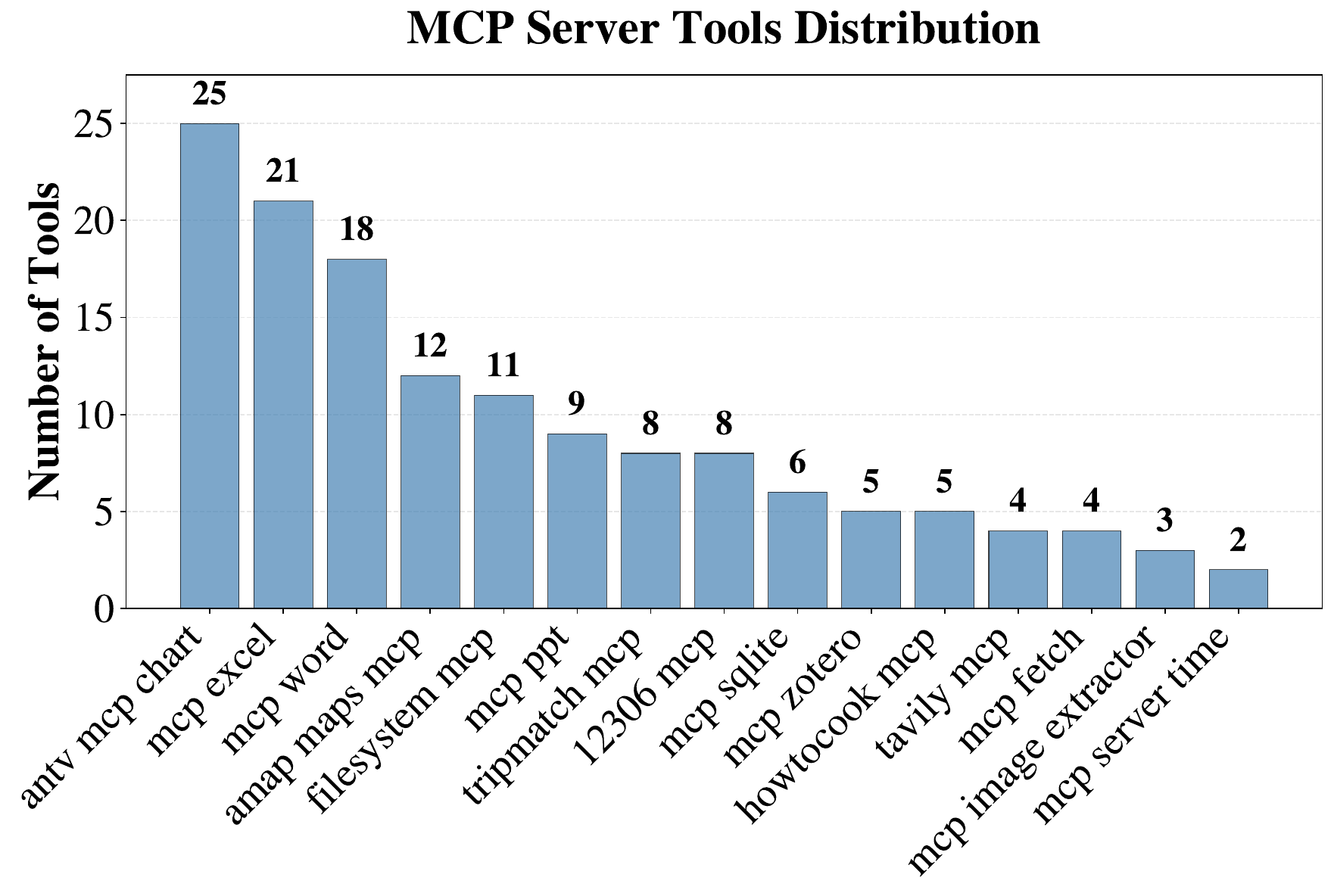}
    \label{fig: subtask+tool: tool}
  \end{subfigure}
  \caption{Statistics of subtasks and tools in TPS-Bench. \textbf{Left:} Distribution of all subtask categories in TPS-Bench. 
  \textbf{Right:} Number of tools in each MCP Server used in TPS-Bench. 
  }
  \label{fig: subtask+tool}
\end{figure}

\section{TPS-Bench}
\label{sec:TPS-Bench}
 
As illustrated in Figure~\ref{fig: TPS-Bench}, the tasks in TPS-Bench are a reasonable composition of multiple subtasks based on a large and diverse tool repository. We present more details below.  



\subsection{Benchmark Construction}
\label{sec:bench-const}

To construct TPS-Bench, we need to first collect a wide range of subtasks. 
Considering the subtasks are actually determined by the tools at hand, we first collect and organize 15 model context protocol (MCP) servers, as shown in \Cref{fig: subtask+tool} (right), each of which provides a set of complementary tools designed to work together. 
These MCP servers provide 141 tools spanning multiple domains.

\begin{figure}[t]
    \begin{center}
        \includegraphics[width=0.9\linewidth]{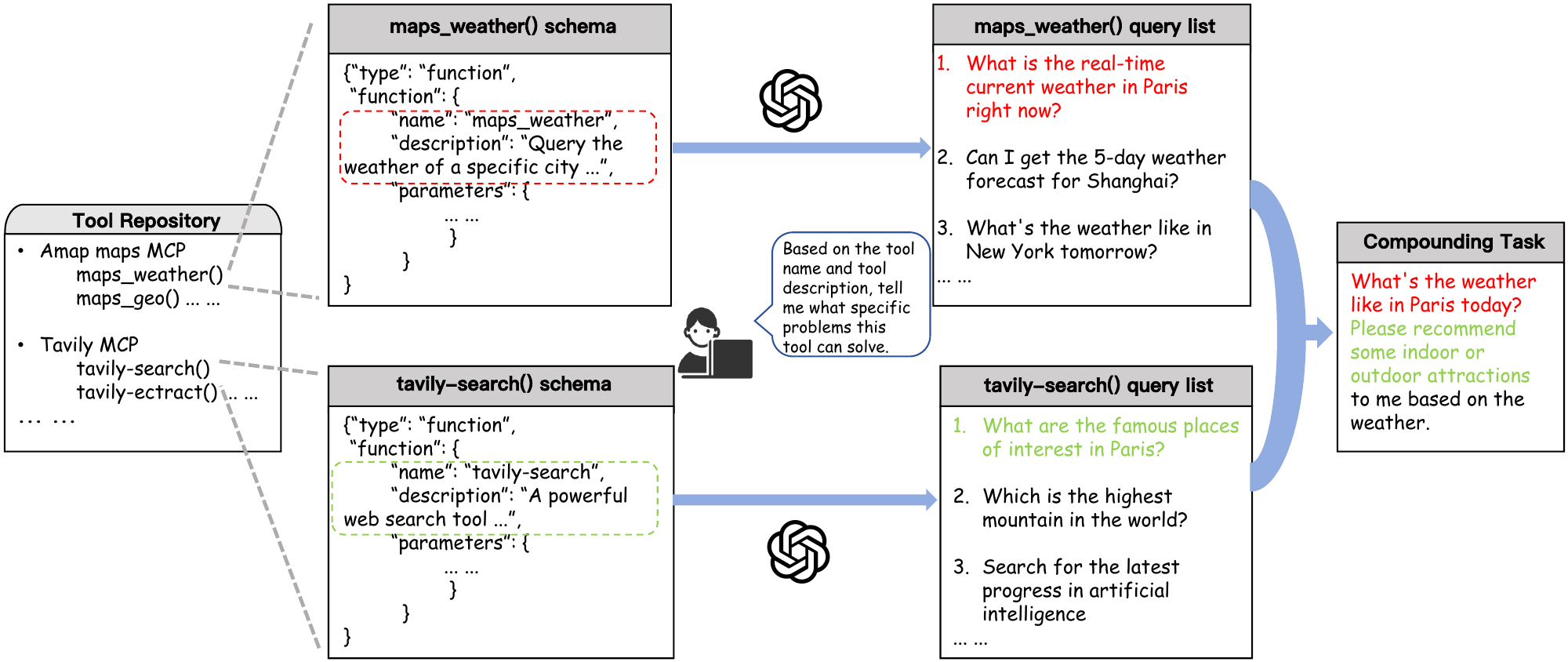}
    \end{center}
    \caption{Task construction workflow in TPS-Bench. We send tool names and descriptions to LLMs to generate solvable subtasks, which are then combined by LLMs and inspected manually to form compounding tasks.}
    \label{fig: Task Construct flow}
\end{figure}

Given such tools, we prompt LLMs to construct the corresponding subtasks that can be solved with the tool repository, as shown in \Cref{fig: Task Construct flow}. 
Specifically, the descriptions of the tools are fed into the LLMs for subtask generation. 
Then, we compose these subtasks, forming two levels of compounding tasks with varying complexity. In detail, the tasks in \textbf{TPS-Bench-Easy} are the combination of simple and weakly relevant subtasks, and the number of subtasks is confined to be at most five. 
For \textbf{TPS-Bench-Hard}, we combine subtasks with strict dependency relationships and 
set the number limit of subtasks to 50. 
The combination is implemented by LLMs and then inspected manually. 


\subsection{Evaluation Process}
\label{sec:eval-process}

To evaluate on TPS-Bench, the LLM agent first needs to perform tool planning---selecting appropriate tools for the task based on the name and description of the tools in our repository. 
We confine the number of selected tools to no more than \textbf{10} to ensure the task completion is efficient in terms of token consumption. 
 Then, the LLM agent needs to decompose the provided task into a set of subtasks and systematically identify the underlying dependencies among them. Based on these dependency relations, the agent determines which tools should be invoked in the current execution turn. 
 We encourage that independent subtasks are executed in parallel to improve efficiency.
 Given the received tool responses, the LLM agent then determines whether there is a need to launch another turn of tool calls. 
 If not, it will summarize the context and return the result to the user.

\subsection{Metrics}
\label{sec:metrics}

\paragraph{Task Completion Rate}
Considering that TPS-Bench tasks are relatively open-ended, we use LLM-as-a-judge~\citep{zheng2023judgingllmasajudgemtbenchchatbot} to evaluate task completion rate. 
To be specific, we introduce a third-party LLM to first decompose the task into a sequence of subtasks, then judge whether each subtask has been completed, and finally calculate the task completion rate and provide a summary. The complete prompt can be found in \Cref{sec:prompt}.
\paragraph{Tool Selection Score}
To assess LLM agents' task planning capabilities, we also evaluate the tool selection score by LLM-as-a-judge. After tool selection, we retrieve the corresponding tool description from the selected tools and submit the query and tool description to LLM for judgment.

We also estimate token and end-to-end time consumption, as well as the number of tool call turns, to measure the efficiency of the scheduling. Regarding time, for open-source LLMs served by ourselves, 
the evaluation is conducted on vLLM~\citep{kwon2023efficient} 
with four NVIDIA A100 GPUs. 
Otherwise, we report the time measured with the official APIs.

\section{Experiment}

The section presents the evaluation results of mainstream LLMs on TPS-Bench. 

\subsection{Setup}
\label{sec:exp-setup}

\paragraph{LLMs}

We consider diverse LLMs in our analysis, including GPT-4o~\citep{openai2024gpt4o}, Kimi-K2~\citep{kimiteam2025kimik2openagentic}, DeepSeek-R1~\citep{deepseekai2025deepseekr1incentivizingreasoningcapability}, GLM-4.5~\citep{zeng2025glm45}, QwQ-32B~\citep{qwq32b}, Qwen3-32B~\citep{qwen3}, and Qwen3-1.7B~\citep{qwen3}.



\paragraph{LLM-as-a-judge}

\begin{figure}[t]
  \centering
  \setlength{\belowcaptionskip}{1pt}
  \begin{subfigure}{0.45\textwidth} 
    \centering
    \includegraphics[width=\linewidth]{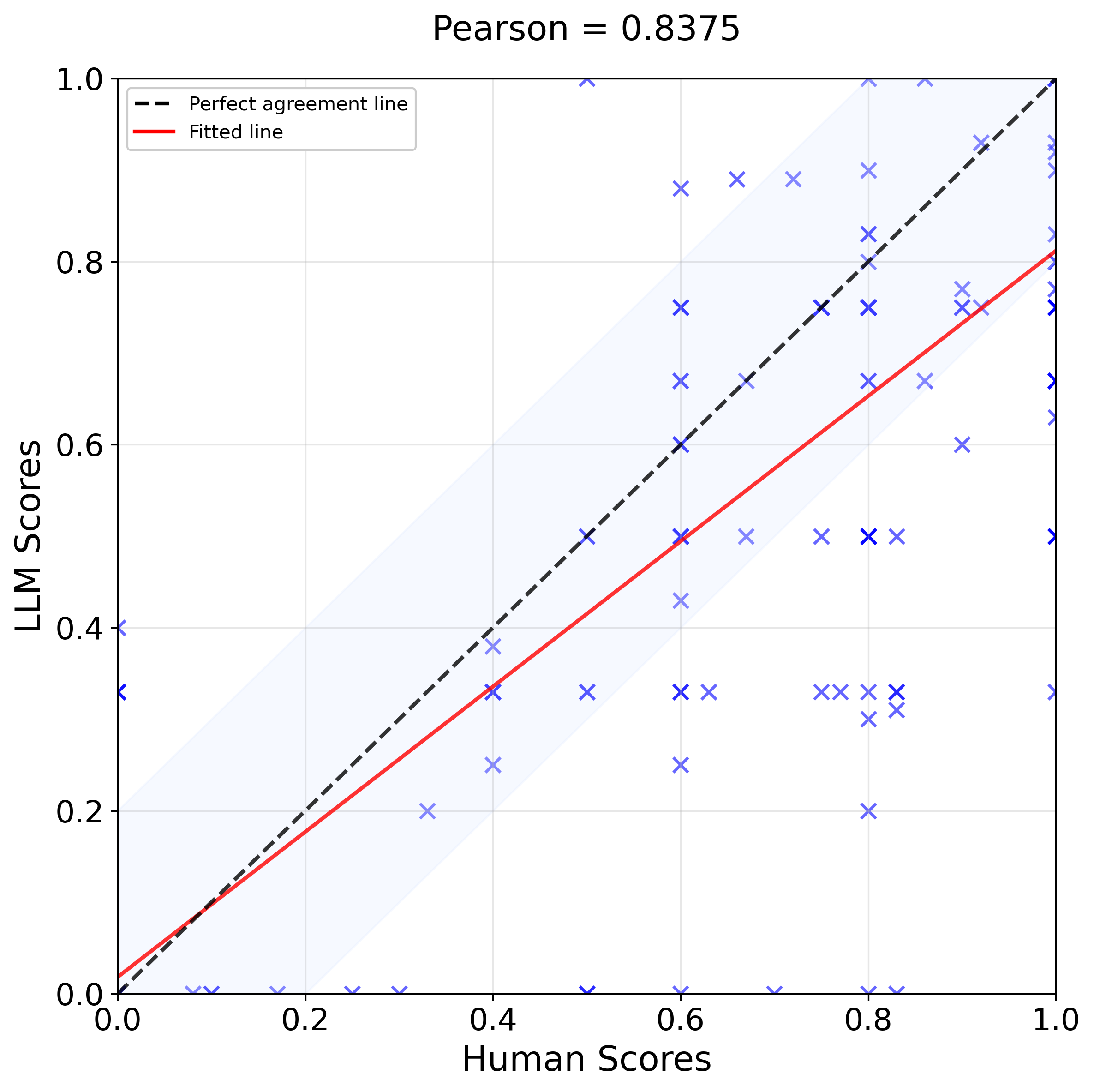}
    \label{fig: Pearson: Scores}
  \end{subfigure}
  \hspace{0.05\textwidth} 
  \begin{subfigure}{0.45\textwidth} 
    \centering
    \includegraphics[width=\linewidth]{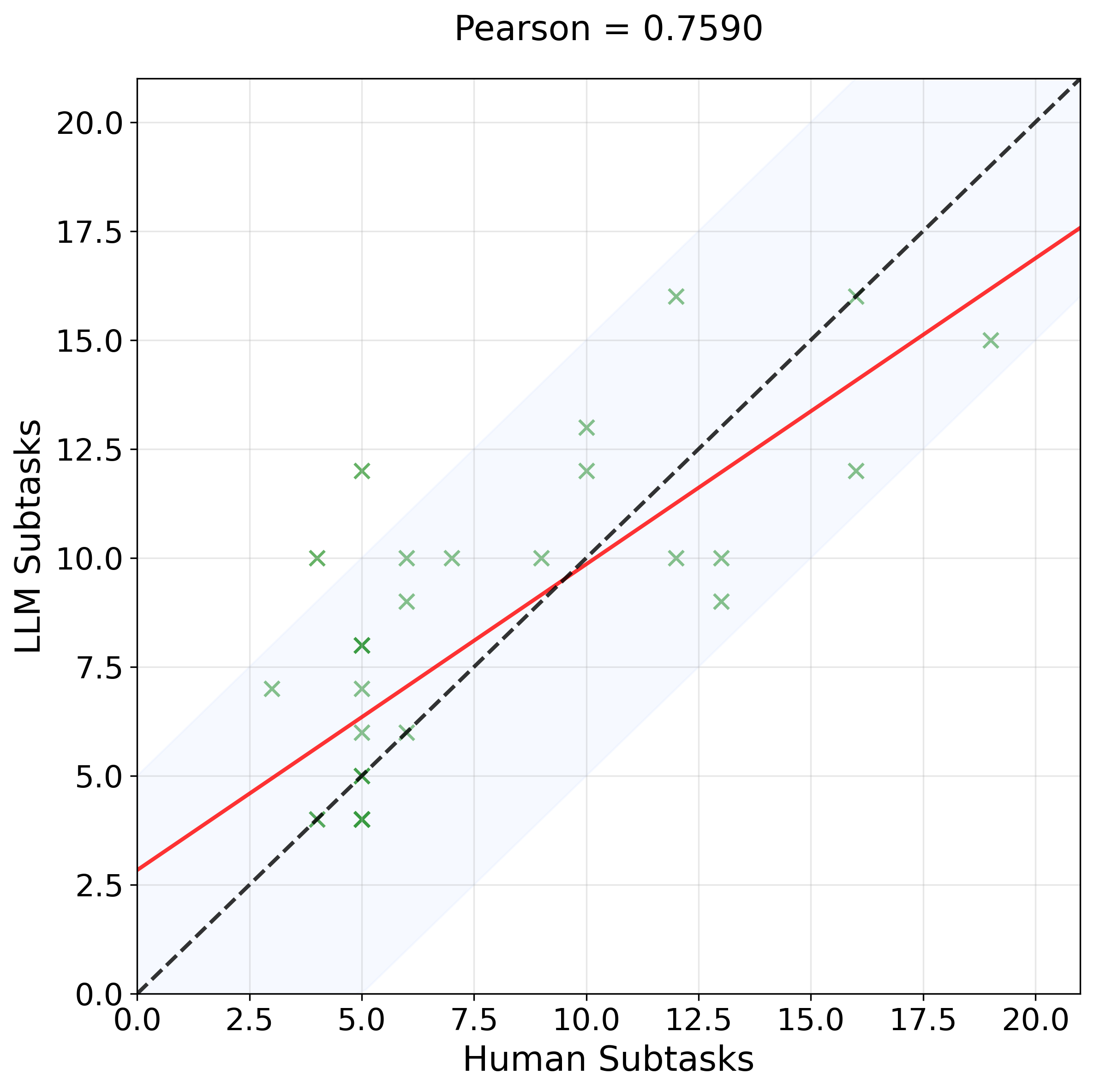}
    \label{fig: Pearson: Subtasks}
  \end{subfigure}
  \caption{\textbf{Left: } Pearson correlation between LLM scores and Human scores on Task Completion Rate. The fitted line (red) shows the linear regression trend, while the dashed black line indicates great agreement between LLMs and humans. \textbf{Right: }Pearson correlation between LLM and Humans regarding the number of subtasks decomposed from the same task.}
  \label{fig: Pearson}
\end{figure}

We employ Gemini-2.5-Flash~\citep{comanici2025gemini25} as the LLM-as-a-judge to evaluate both the task completion rate and tool selection score.
To verify its reliability, we provide the correlation between human evaluations and the LLM-as-a-judge in Figure~\ref{fig: Pearson}.
As shown, the Pearson correlation coefficient is 0.8375. 
We also measure the correlation between the number of subtasks decomposed by the LLM and by humans for the same tasks, and observe a Pearson coefficient of 0.7590. The strong positive correlations (0.8375 and 0.7590) demonstrate that LLM performance is closely aligned with human judgments. This indicates that TPS-Bench provides a reliable benchmark for evaluating both task success and decomposition quality.



\begin{table}[htbp]
  \scriptsize  
  \setlength{\tabcolsep}{2pt} 
  \renewcommand{\arraystretch}{1.15}
  \centering
  
  \begin{tabular}{c c c c c c c}  
    \toprule
    \multirow{2}{*}{Model} & \multicolumn{2}{c}{Effectiveness} & \multicolumn{4}{c}{Efficiency} \\
    \cmidrule(lr){2-3} \cmidrule(lr){4-7}  
    & Tool Selection Score$\uparrow$ & Task Completion Rate$\uparrow$ & \# Input Tokens$\downarrow$ & \# Output Tokens$\downarrow$ & Turns$\downarrow$ & Time$\downarrow$ \\ 
    \midrule
    
    \multicolumn{7}{c}{\textbf{TPS-Bench-Easy}} \\  
    \midrule
    
    GPT-4o~\citep{openai2024gpt4o}          & 94.09\% & 52.44\% & 7.2k & 0.8k  & 2.2 & 51.2 \\
    Kimi-K2~\citep{kimiteam2025kimik2openagentic}         & 92.87\% & 73.54\% & 6.5k & 0.8k & 3.7 & 103.2 \\
    DeepSeek-R1~\citep{deepseekai2025deepseekr1incentivizingreasoningcapability}     & 92.19\% & 68.52\% & 6.5k & 1.1k & 2.8 & 329.0 \\
    GLM-4.5~\citep{zeng2025glm45}         & 87.72\% & 64.91\% & 11.1k & 1.2k & 9.5 & 123.0 \\
    QwQ-32B~\citep{qwq32b}         & 85.36\% & 32.65\% & 6.5k & 1.1k & 1.0 & 217.6 \\
    Qwen3-32B~\citep{qwen3}       & 93.03\% & 61.60\% & 6.3k & 0.9k & 2.5 & 202.3 \\
    Qwen3-1.7B~\citep{qwen3}      & 78.03\% & 33.35\% & 7.3k & 1.1k & 2.3 & 22.2 \\
    \midrule
    
    \multicolumn{7}{c}{\textbf{TPS-Bench-Hard}} \\ 
    \midrule

    GPT-4o~\citep{openai2024gpt4o}          & 87.12\% & 45.08\% & 7.2k & 1.3k & 2.5 & 76.84 \\
    Kimi-K2~\citep{kimiteam2025kimik2openagentic}         & 84.49\% & 52.29\% & 6.5k & 1.1k & 19.7 & 216.5 \\
    DeepSeek-R1~\citep{deepseekai2025deepseekr1incentivizingreasoningcapability}     & 84.41\% & 62.03\% & 6.7k & 1.4k & 6.0 & 343.4 \\
    GLM-4.5~\citep{zeng2025glm45}         & 79.31\% & 64.72\% & 12.6k & 1.5k & 35.0 & 217.8 \\
    QwQ-32B~\citep{qwq32b}         & 70.74\% & 29.36\% & 6.7k & 1.6k & 1.05 & 171.0 \\
    Qwen3-32B~\citep{qwen3}       & 85.36\% & 56.72\% & 6.7k & 1.3k & 3.1 & 226.2 \\
    Qwen3-1.7B~\citep{qwen3}      & 65.26\% & 26.75\% & 7.8k & 2.2k & 2.4 & 42.0 \\
    \bottomrule
  \end{tabular}

  \caption{
  The main results of testing 7 representative models on TPS-Bench are presented here. We report tool selection score and task completion rate (higher is better), and token usage (Input Tokens, Output Tokens; lower is better). Regarding execution time, for the Qwen3-1.7B model, the evaluation was conducted locally based on vLLM, whereas for all other models, the time was obtained via API calls in their default settings.
  }
  \label{tab:main result}
\end{table}

\subsection{Main Results}
\label{sec:exp-main}


\paragraph{Reasoning and Planning Abilities Count}


Table~\ref{tab:main result} reports the tool selection score and task completion rate of 7 representative models.
The results show that large reasoning models (LRMs) like Kimi-K2, DeepSeek-R1, and GLM-4.5 achieve a task completion rate of 52.29\%, 62.03\%, and 64.72\%, respectively, which have comprehensively outperformed GPT-4o. Among them, GLM-4.5 has maintained the highest completion rate of 64.72\% on TPS-Bench-Hard, demonstrating both robust reasoning and consistent decision-making abilities across diverse tasks. However, LRMs clearly consume significantly more tokens and time—for instance, DeepSeek-R1 takes an average of 343s and 8k tokens to complete a single task. It reflects the performance differences between models with varying planning and reasoning capabilities on compounding tasks. Furthermore, we found that although Qwen3-32B has a much smaller parameter size than Kimi-K2 and GPT-4o, it still maintains competent performance in the task completion rate of 56.72\%. Such efficiency, coupled with its relatively lightweight architecture, may be attributed to its training on agentic data.

\paragraph{Different Scheduling Strategies Affect Efficiency}

Table~\ref{tab:main result} indicates that most models exhibit reasonable performance in tool planning but behave differently in scheduling strategies. 
In particular, GLM-4.5 requires an average of up to 35 tool call turns per task on TPS-Bench-Hard.
As shown in Table 1, GLM-4.5 performs a considerable number of tool call turns under both difficulty levels. We note that GLM-4.5 tends to perform tool execution in a strictly sequential manner, and it achieves a task completion rate as high as 64.72\%. However, GLM-4.5 executes an average of 35 tool call turns per task, takes 217.8 seconds, and uses as high as 12.6k input tokens, exhibiting poor efficiency. In contrast, GPT-4o tends to execute tool execution in parallel, with an average of 2 tool call turns per task. While the time required is significantly reduced, its task completion rate is only 45.08\%. As for QwQ-32B, it struggles to clearly distinguish dependencies among subtasks and often invokes multiple tools simultaneously, even when certain tools rely on the outputs of others. This behavior results in only one tool call turn being produced and leads to the low task completion rate of 29.36\%.

\paragraph{Cost}

To better capture the trade-off between monetary cost and task performance, we further collect and report the cost-of-pass for each model in Table~\ref{tab:main cost}. The cost-of-pass denoted as $v(m, t)$, represents the expected monetary cost of employing a model $m$ to complete a task $t$. This metric allows a fair comparison between models with distinct pricing structures and completion capabilities. It is formulated as the ratio between the overall cost of completing a task once, $C_m(t)$, and the corresponding task completion rate, $R_m(t)$:
\begin{equation}
    v(m,t) = \frac{C_m(t)}{R_m(t)} .
\end{equation}

The total cost of completing a task once, $C_m(t)$, is given by:
\begin{equation}
    C_m(t) = n_{\text{in}}(m,t) \cdot c_{\text{in}}(m) + n_{\text{out}}(m,t) \cdot c_{\text{out}}(m) ,
\end{equation}
where $n_{\text{in}}(m,t)$ and $n_{\text{out}}(m,t)$ denote the number of input and output tokens consumed by model $m$ on task $t$, while $c_{\text{in}}(m)$ and $c_{\text{out}}(m)$ are their respective per-token costs. 

As the table, GPT-4o incurred the highest cost, reaching 52.2 $\times$ $10^{-3}\$$, and the highest cost-of-pass, reaching 138.0 $\times$ $10^{-3}\$$, primarily due to its premium pricing. In contrast, QwQ-32B required a low cost, yet its cost-of-pass was six times higher because of its low task completion rate on TPS-Bench.

\begin{table}[t]
  \small
  \setlength{\tabcolsep}{8pt}
  \renewcommand{\arraystretch}{1.15}
  \centering
  \begin{tabular}{c *{2}{c} *{2}{c}}  
    \toprule
    \multirow{2}{*}{Model} & \multicolumn{2}{c}{TPS-Bench-Easy} & \multicolumn{2}{c}{TPS-Bench-Hard} \\
    \cmidrule(lr){2-3} \cmidrule(lr){4-5} 
    & Cost$\downarrow$ & Cost-of-Pass$\downarrow$ & Cost$\downarrow$ & Cost-of-Pass$\downarrow$ \\
    \midrule
    GPT-4o          & 52.6 & 100.3 & 62.2 & 138.0 \\
    Kimi-K2         & 5.9 & 8.0 & 6.5 & 12.4 \\
    DeepSeek-R1     & 5.8 & 8.5 & 6.7 & 10.8 \\
    GLM-4.5         & 9.8 & 8.5 & 11.3 & 17.5 \\
    QwQ-32B         & 2.7 & 15.3 & 3.2 & 21.2 \\
    Qwen3-32B       & 4.3 & 7.0 & 5.5 & 9.7 \\
    Qwen3-1.7B      & 0.8 & 2.4 & 1.3 & 4.9 \\
    \bottomrule
  \end{tabular}
  \vspace{6pt}
  \caption{The average monetary cost of the seven models on TPS-Bench, including the average cost($10^{-3}\$$) and cost-of-pass for tasks at each level. More information about the prices of each model can be found in Appendix~\ref{sec:modelprice}.}
  \label{tab:main cost}
\end{table}

\subsection{Discuss and Ablation}
\label{sec:exp-ablation}
The strong performance of our models on TPS-Bench primarily depends on two factors: the ability to select appropriate tools and the ability to schedule them correctly in sequential and parallel workflows. To gain a deeper understanding of how each of these factors affects the model’s tool planning and scheduling capabilities, we conducted two corresponding ablation studies.

\paragraph{The Impact of Tool Selection Strategy}
In previous experiments, we all had the model select tools on its own, and we refer to this strategy as ``self-selection''.
To investigate the impact of different tool selection strategies, we designed two alternative tool selection strategies for comparison.

\begin{itemize}
\item \textbf{No selection (all tools used) : }In this setting, no tool selection is performed before task execution; instead, all available tool schemas are provided to the model directly.

\item \textbf{Rule-based selection: }In this setting, the task content is compared with the tool names and descriptions using word-based similarity measures. The top-10 most similar tools are then selected and provided to the model for task execution.
\end{itemize}

\begin{figure}[h]
    \begin{center}
        \includegraphics[width=0.9\textwidth]{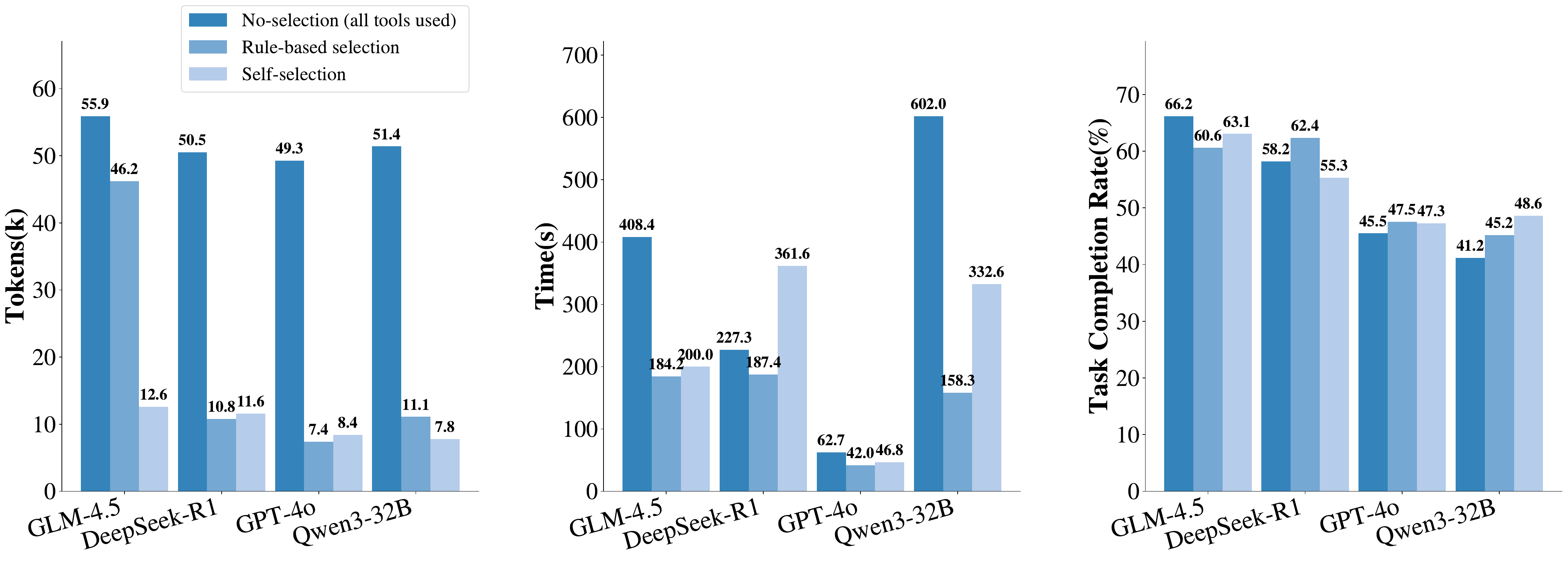}
    \end{center}
    \caption{Evaluation of three tool selection strategies, no-selection, rule-based selection, and self-selection, was performed by applying each strategy to the tasks in TPS-Bench-Hard. The four models, GLM-4.5, DeepSeek-R1, GPT-4o, and Qwen3-32B, are tested with all three strategies across the benchmark. Efficiency is reflected by the total number of tokens consumed and the time required to complete each task, while effectiveness is reflected by the task completion rate.}
    \label{fig: ablation tool}
\end{figure}

For this ablation experiment, we employed GLM-4.5, DeepSeek-R1, GPT-4o, and Qwen3-32B. The results are shown in Figure~\ref{fig: ablation tool}. It indicates that different tool selection strategies do not lead to substantial differences in task completion rate. However, notable differences are observed in the total number of tokens consumed and the time required. The no-selection strategy, which inputs a large number of tool schemas, results in a token count significantly higher than the other two strategies, exceeding 50k tokens. Additionally, since the model needs to select tools on its own, both the no-selection and self-selection strategies require more time. These findings indicate that although the task completion rate is largely unaffected by the choice of tool selection strategy, the efficiency of each strategy varies significantly, emphasizing the importance of appropriate tool selection.

\begin{wrapfigure}[16]{r}{0.4\textwidth}  
    \centering  
    \includegraphics[width=0.4\textwidth]{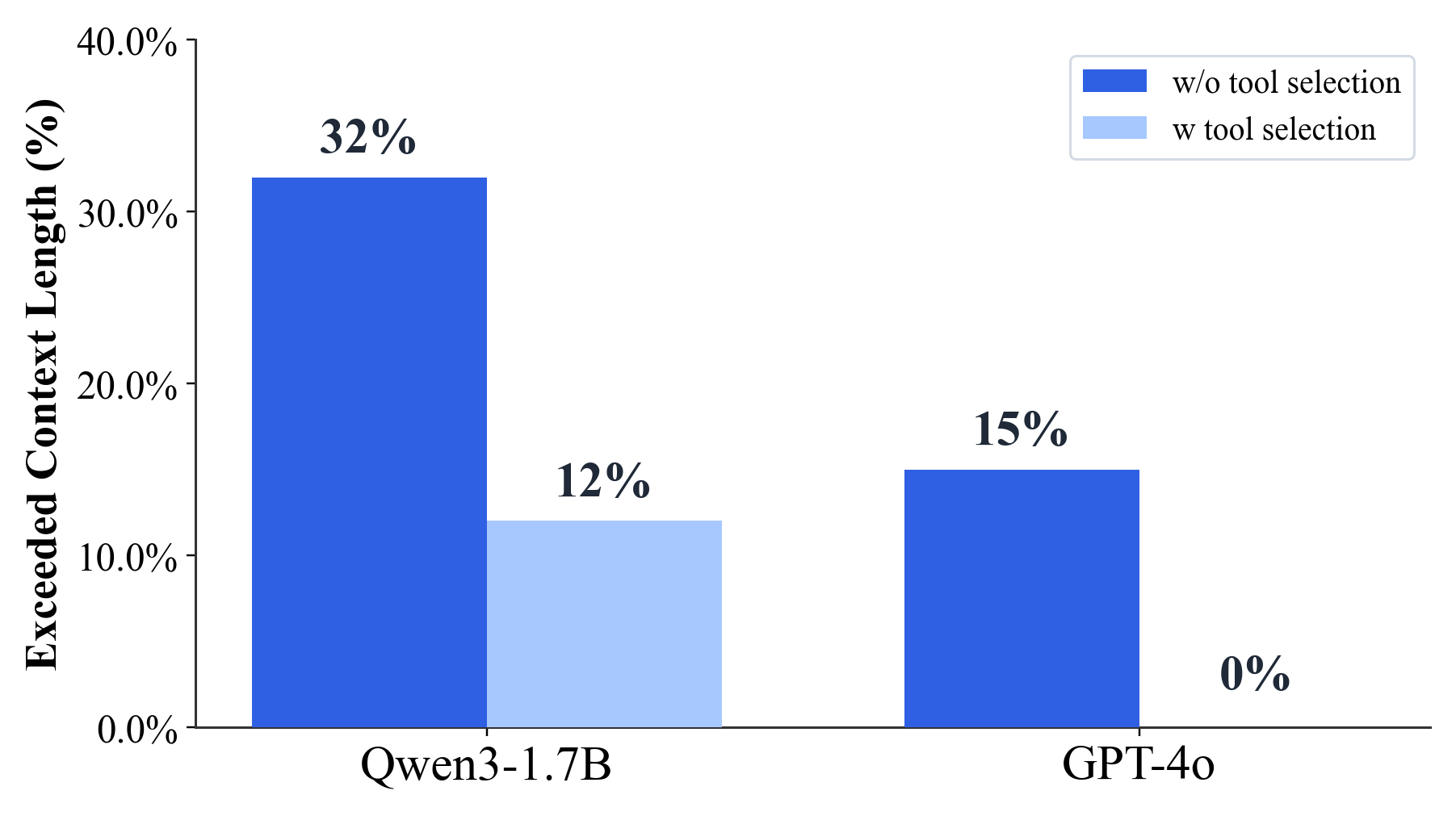}  
    \caption{Evaluation of the impact of tool selection on context length exceeding for Qwen3-1.7B and GPT-4o. The tool selection strategy reduces the occurrence of context length exceeding caused by a long tool schema.}
    \label{fig: context limit}
\end{wrapfigure}

In addition, for small models with a relatively short context length—such as Qwen3-1.7B, which has a 40k context length—it is prone to exceeding the context length. As shown in the Figure~\ref{fig: context limit}, without tool selection, 32\% of cases of Qwen3-1.7B exceeded the context length, and this proportion dropped to 12\% after tool selection. For GPT-4o, the proportion of these cases also decreased from 12\% to 0\% after tool selection.

\begin{figure}[t]
    \begin{center}
        \includegraphics[width=0.9\textwidth]{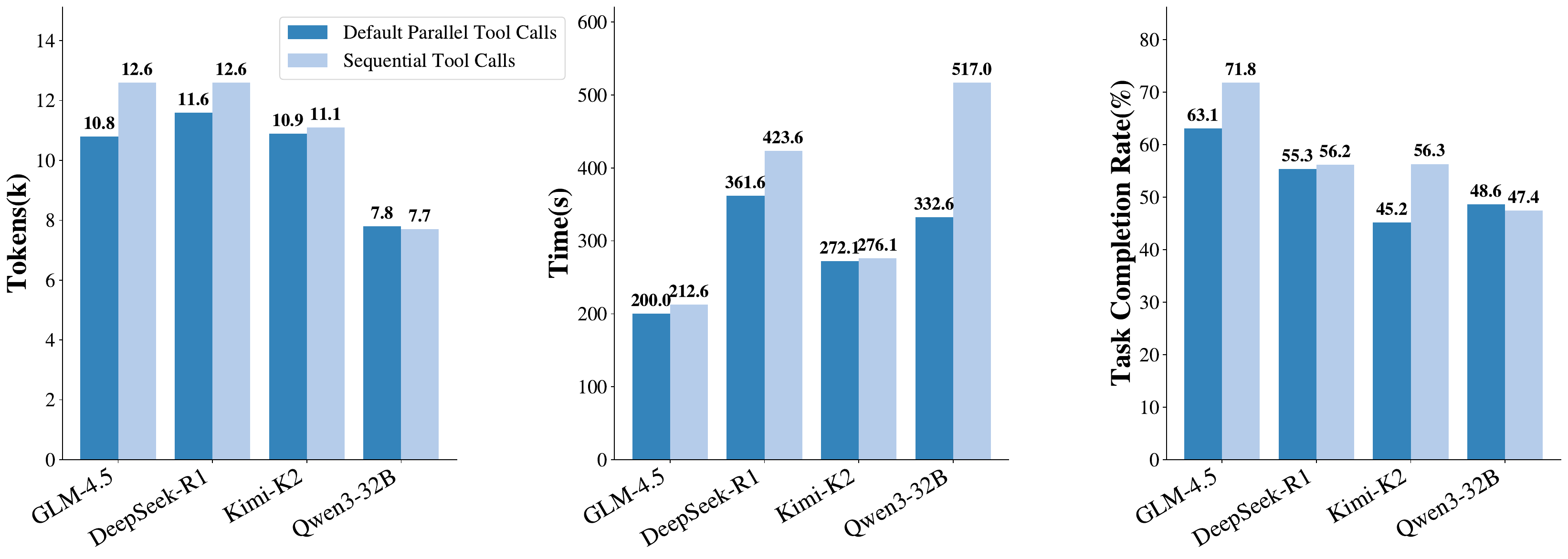}
    \end{center}
    \caption{Evaluation of two tool scheduling strategies, default parallel tool calls and sequential tool calls, was conducted on TPS-Bench-Hard. We tested four models, GLM-4.5, DeepSeek-R1, Kimi-K2, and Qwen3-32B, applying both scheduling strategies across the benchmark to ensure comparability. Differences between the two strategies are reflected in the number of tokens consumed, the time required, and the task completion rate, thereby illustrating the trade-offs between efficiency and effectiveness when tools are invoked in parallel versus sequentially.}
    \label{fig: ablation para ser}
\end{figure}

\paragraph{Tool Scheduling}
To compare the advantages and disadvantages of serial and parallel tool scheduling in terms of both effectiveness and efficiency, we conducted experiments using four models. As Figure~\ref{fig: ablation para ser}, serial scheduling strategies generally consume more tokens and time than parallel strategies. For instance, DeepSeek-R1’s token usage increased from 11.6k to 12.6k, and time consumption rose from 361.6 seconds to 423.6 seconds. During serial execution, the model engages in additional reasoning steps, which incur extra token and time costs. These extra steps allow the model to evaluate subtask dependencies, potentially avoiding cascading errors. Thereby, the task completion rate is also somewhat improved, for example, GLM-4.5 increased from 63.1\% to 71.8\%. Serial scheduling can effectively help the model reduce errors caused by incorrect judgments of subtask dependencies. These results highlight a clear trade-off between efficiency and effectiveness, suggesting that the choice of scheduling strategy should consider both computational cost and task reliability.

\section{Reinforcement Learning}

To improve the scheduling ability without compromising performance, especially in complex and multi-tool scenarios, 
considering the trade-off between model performance during sequential tool calls and efficiency in parallel execution, this section presents an initial study that uses Qwen3-1.7B~\citep{qwen3} to enhance scheduling efficiency through RL training.

\paragraph{Training Details} We utilize Group Relative Policy Optimization (GRPO)~\citep{shao2024deepseekmath} to train Qwen3-1.7B on 100 training samples over five epochs with an actor learning rate of 1e-6. For each training step, 5 roll-out completions are generated for each sample, with all other hyperparameters set to the default values provided by veRL~\citep{sheng2024hybridflow}. The reward signal is derived from Gemini-2.5 flash~\citep{comanici2025gemini25}, which scores the model on task completion and the degree of parallelism. This specific reward design directly incentivizes the model to balance effectiveness and efficiency.

\paragraph{Result Analysis} As shown in \Cref{tab:grpo}, with only 100 training samples, the GRPO-trained Qwen3-1.7B model demonstrates remarkable   improvements across key metrics on the TPS-Bench-Hard. This is evidenced as the model improves the average task completion rate by 6\% while simultaneously reduces execution time by 14\% in TPS-Bench-Hard. Notably, compared with the baseline, the model requires fewer tool call turns and generates fewer output tokens, which indicates more parallel tool execution, suggesting that reinforcement learning can effectively guide the model toward more structured and parallel execution patterns.


\begin{table}[htbp]
  \small 
  \setlength{\tabcolsep}{3pt}  
  
  \centering
    \begin{tabular}{c c *{6}{c}}  
    \toprule
    \multirow{2}{*}{Model} & \multirow{2}{*}{Method} 
    & \multicolumn{6}{c}{TPS-Bench-Hard} \\
    \cmidrule(lr){3-8}  
    &  & Tool Sel. Score & Task Comp. Rate 
       & \# Input Tokens & \# Output Tokens 
       & Turn & Time \\
    \midrule
    \multirow{2}{*}{Qwen3-1.7B} 
    & origin   & 65.26\% & 26.75\% & 7.8k & 2.2k & 2.4 & 42.0 \\
    & GRPO     & 81.18\% & 33.13\% & 7.3k & 1.0k & 2.1 & 36.1 \\
    \bottomrule
  \end{tabular}
  \vspace{6pt}
  \caption{Results of Qwen3-1.7B on TPS-Bench-Hard after tuning on 100 RL samples}
  \label{tab:grpo}
\end{table}

\section{Conclusion}
We propose TPS-Bench to evaluate models’ tool planning and scheduling capabilities on real-world compounding tasks. To better showcase these capabilities, we build an agent system that allows the model to self-select tools before task execution, and construct a tool pool containing hundreds of tools. Model performance is then evaluated from both effectiveness and efficiency perspectives.

Our experiments reveal that mainstream models still struggle with insufficient task completion and low efficiency when tackling compounding tasks, reflecting their limitations in tool planning and scheduling. To validate the impact of tool selection and scheduling strategies, we conducted corresponding ablation studies, confirming that selecting appropriate tools and leveraging parallel execution of subtasks can significantly improve both effectiveness and efficiency. Finally, by training Qwen3-1.7B with GRPO, we observed a 14\% reduction in execution time alongside a 6\% improvement in task completion rate.

Through TPS-Bench, our goal is to evaluate and advance the tool planning and scheduling capabilities of LLM agents, enabling them to address real-world compounding problems both effectively and efficiently. While our benchmark cannot fully capture every aspect of real-world scenarios, we believe it represents a meaningful step toward enhancing the capabilities of LLM agents.







\bibliography{iclr2026_conference}
\bibliographystyle{iclr2026_conference}

\appendix

\section{Statement on the Use of LLM Assistance}
\label{sec:appendix-llm}

Consistent with community guidelines on responsible use of large language models (LLMs), we disclose that LLM tools were used only to assist with language editing (grammar, wording, and minor style) of this manuscript. All ideas, claims, methods, experiments, analyses, figures, and tables were conceived, implemented, and verified by the authors. The authors reviewed and edited all LLM-suggested text for accuracy and clarity; no passages were accepted without human verification. LLMs were not used to generate data, code, results, reviews, or citations, and no confidential or proprietary information was provided to LLM services.

\section{PROMPT FOR TPS-Bench}
\label{sec:prompt}

\subsection{Prompt for Tool Scheduling }
\begin{tcolorbox}[
    colback=gray!15,    
    colframe=gray!30,   
    boxsep=8pt,         
    sharp corners,
    breakable
]
"""

   You are a helpful assistant specialized in efficient task execution and parallel processing.

   $\#$ Tool Usage Instructions \\
   If you need to search for something, please set max-result==5. \\

   $\#$ Task Execution Framework\\
   
   When solving a user's problem, please strictly follow these steps: \\

   $\#\#$ 1. Task Analysis and Decomposition\\
   - Break down the user's question into specific, manageable sub-tasks\\
   - Each sub-task should have a single, clear objective\\
   - Ensure sub-tasks are granular enough to be independently executed and verified\\
   - Consider different decomposition strategies: functional modules, data flow, temporal sequence\\

   $\#\#$ 2. Dependency Analysis\\
   For each sub-task, identify:\\
   - **Data dependencies**: Tasks requiring output from other tasks\\
   - **Logical dependencies**: Tasks with sequential execution requirements\\
   - **Resource dependencies**: Tasks competing for the same resources\\
   - Create a dependency graph to visualize relationships\\

   $\#\#$ 3. Execution Planning\\
   - **Sequential tasks**: Must be executed in order due to dependencies\\
   - **Parallel tasks**: Can be executed simultaneously with resource considerations\\
   - **Priority assignment**: Critical path tasks get higher priority\\
   - **Resource allocation**: Balance CPU, memory, and network usage\\
   - **Error handling**: Define fallback strategies for failed parallel tasks\\

   $\#\#$ 4. Execution Monitoring\\
   - Track progress of each sub-task\\
   - Monitor resource usage and performance\\
   - Implement timeout mechanisms\\
   - Prepare for dynamic plan adjustments\\

   $\#\#$ 5. Quality Assurance\\
   - Verify results of each sub-task\\
   - Ensure integration of parallel results is coherent\\
   - Validate final output meets user requirements\\

   $\#\#$ 6. Task Completion Summary\\
   - When you complete tasks, provide detailed descriptions of each sub-task performed\\
   - Explain the specific actions taken and results achieved for every sub-task\\
   - Include any challenges encountered and how they were resolved\\
   - **IMPORTANT: Always provide your complete answers and solutions in the final response, regardless of whether you save them to docx, pptx, or other files. Users should be able to understand your complete solution process and results without needing to check external files**\\
   - Summarize the overall completion status and final outcomes\\

   $\#\#$ CRITICAL EXECUTION REQUIREMENTS:\\
   - **SILENT EXECUTION**: Do NOT output intermediate steps, progress updates, or ask "what should I do next?"\\
   - **COMPLETE ALL WORK**: Execute all necessary sub-tasks in one continuous session without stopping\\
   - **FINAL SUMMARY ONLY**: Only provide the comprehensive final summary with all results, data, and conclusions\\
   - **NO INTERRUPTIONS**: Never pause to ask for user input or next steps - complete everything autonomously\\
   - **COMPREHENSIVE OUTPUT**: Include all research findings, analysis, data, and conclusions in your final response\\

   Execute all tasks silently and provide only the complete final summary with all results.

   Before each operation, explain your plan and reasoning in natural language, then proceed to call the tool(s). You may call multiple tools in parallel for independent tasks, but always consider resource constraints and error handling.
   
   """

\end{tcolorbox}

\subsection{Prompt for Tool Selection}
\begin{tcolorbox}[
    colback=gray!15,    
    colframe=gray!30,   
    boxsep=8pt,         
    sharp corners,
    breakable
]

"""
\\
You are a tool selection expert. Please select the most relevant tools from the provided tool list based on the user's query. \\

User query: {query} \\
IMPORTANT: \\
- Select only the tools that are necessary and relevant to complete the task\\
- Maximum 10 tools, but you can select fewer if appropriate\\
- Output format must be exactly like this (one tool name per line):\\
\verb|tool_name_1|\\
\verb|tool_name_2|\\
\verb|tool_name_3|\\
\verb|...|\\
No explanations, no analysis, no additional text.\\
"""

\end{tcolorbox}

\subsection{Prompt for Evaluating Task Completion Rate}
\begin{tcolorbox}[
    colback=gray!15,    
    colframe=gray!30,   
    boxsep=8pt,         
    sharp corners,
    breakable
]
"""
    You are a professional evaluation assistant. Your task is to assess the task completion rate of AI models. \\

    \#\# Task Context\\
    {query}\\

    \#\# Instructions\\

    \#\#\# Step 1: Task Decomposition\\
    Based on the task context above, break down the overall task into granular, indivisible subtasks. Each subtask should be:\\
    - Specific and measurable\\
    - Independently verifiable\\
    - Cannot be further subdivided\\

    **Examples:**\\
    
    Example 1: "I need help with several things: search for Tesla's latest earnings report and create a presentation, convert the address 'Empire State Building, New York' to coordinates, get the current time in Tokyo, and find some healthy breakfast recipes for weight loss."\\
    Total subtasks: 5\\
    1. Search for Tesla's latest earnings report\\
    2. Create a presentation (based on the earnings report)\\
    3. Convert the 'Empire State Building, New York' address to coordinates\\
    4. Get the current time in Tokyo\\
    5. Find healthy breakfast recipes for weight loss\\
    
    Example 2: "My family wants to buy properties for investment in several cities and needs a comprehensive market analysis. The main cities under consideration are Chengdu, Wuhan, and Nanjing. I want to understand housing price trends, surrounding facilities, and transportation convenience in these places. I'm particularly concerned about the distribution of schools, hospitals, and shopping centers, as well as commuting time to city centers. All analysis results need to be organized into a detailed investment report with comparative data and visualization charts. Additionally, I want to understand the recent weather and environmental conditions in these cities, as living comfort is also important. The final report should clearly show the advantages and disadvantages of each city to help make the best investment decision."\\
    Total subtasks: 23 (3 cities * 7 analysis aspects + 2 report tasks)\\
    For each city (Chengdu, Wuhan, Nanjing):\\
    1. Analyze housing price trends\\
    2. Evaluate surrounding facilities\\
    3. Assess transportation convenience\\
    4. Map distribution of schools\\
    5. Map distribution of hospitals\\
    6. Map distribution of shopping centers\\
    7. Calculate commuting time to the city center\\
    Additional tasks:\\
    22. Research recent weather and environmental conditions for all cities\\
    23. Create a comprehensive investment report with comparative data and visualization charts\\

    Please list all subtasks clearly with numbers (1, 2, 3, etc.)\\

    \#\#\# Step 2: Model Response Analysis\\
    Model Response:\\
    \{result\}\\

    **Special Case: Empty Response**\\
    If the model response (result) is empty, null, or contains no meaningful content, then all subtasks are incomplete.\\

    **Special Case: Tool Calls**\\
    If the model response has \verb|<tool_call>| tags, then the relative subtasks are incomplete.\\

    **Special Case: Documents**\\
    If the task requires creating or providing a document, then the response should include the creation of files in the ".docx" or ".pptx" format.\\
    - If the model response mentions creating visualization files, like"xxx.docx" or "xxx.pptx", and provides relevant conclusions or insights based on those visualizations, then the corresponding subtask should be considered completed even if the actual file is not accessible for verification. \\
    - If the response does not explicitly mention the specific file names of the created files, like "xxx.docx" or "xxx.pptx", then not only the file creation subtasks but also the subtasks related to adding content to the files should be considered incomplete.\\

    \#\#\# Step 3: Completion Assessment\\
    For each subtask you identified in Step 1, analyze whether it was completed in the model response:\\
    -  Completed successfully: Provide evidence from the response\\
    -  Not completed: Explain why it wasn't addressed\\
    -  Partially completed: Explain what was done and what was missing\\

    \#\#\# Step 4: Final Scoring\\
    Calculate the completion percentage based on fully completed subtasks only.\\

    \#\#Output Format\\
    Please provide a detailed analysis in the following format:\\

    **SUBTASK DECOMPOSITION:**\\
    1. First subtask\\
    2. Second subtask\\
    ...\\
    Total subtasks: X\\

    **COMPLETION ANALYSIS:**\\
    1. Subtask 1 - Explanation with evidence\\
    2. Subtask 2 - Explanation with evidence\\
    ...

    **FINAL RESULT:**\\
    - Completed subtasks: X out of Y\\
    - Completion rate: XX\% \\
    - Summary: Brief explanation of overall performance

\end{tcolorbox}

\subsection{Prompt for evaluating Tool Selection Score}
\begin{tcolorbox}[
    colback=gray!15,    
    colframe=gray!30,   
    boxsep=8pt,         
    sharp corners,
    breakable
]

""" \\
    You are a tool selection evaluation expert. Please evaluate the quality of the model's tool selection based on the following criteria:\\

    **Evaluation Task:**\\
    User Query: \\
    \{query\}\\
    Model Selected Tools: \\
    \{selected tools\}\\
    Available Tools List: \\
    \{available tools\}\\

    **Evaluation Criteria:**\\

    1. **Tool Matching Assessment (50 points)**\\
    - Do the selected tools directly address the specific needs in the user query?\\
    - Do the selected tool functions match the task requirements?
    - Are there any obvious missing essential tools?\\
    - Is the logic of tool selection reasonable?\\

    2. **Efficiency Assessment (30 points)**\\
    - Were redundant or duplicate-function tools selected?\\
    - Is the number of tools reasonable (maximum 10)?\\
    - Is there a more concise tool combination solution?\\

    3. **Completeness Assessment (20 points)**\\
    - Does it cover all subtasks in the user query?\\
    - Can the tool combination form a complete solution?\\

    **Scoring Rules:**\\
    - Total score: 100 points, allocated according to the above weights\\
    - Exceeding 10 tools: deduct 5 points for each additional tool\\
    - Missing critical tools: deduct 10 points for each missing tool\\
    - Selecting irrelevant tools: deduct 8 points for each irrelevant tool\\

    **Output Format:**\\
    {{\\
    "total score": X,\\
    }}\\

    Please conduct the evaluation strictly according to the above criteria and provide an objective, detailed analysis.\\

"""
\end{tcolorbox}

\clearpage

\section{PRICE OF MODELS}
\label{sec:modelprice}

\begin{table}[htbp]
  \centering
  \caption{Pricing of Each Model per 1M Tokens of \textbf{July 30, 2025}  (Unit: $\$$)}
  \begin{tabular}{|c|c|c|}
    \hline
    \textbf{Model Name} & \textbf{Input Pricing (per 1M tokens)} & \textbf{Output Pricing (per 1M tokens)} \\
    \hline
    GPT-4o & 5.00 & 20.00 \\
    \hline
    Kimi-K2 & 0.60 & 2.50 \\
    \hline
    DeepSeek-R1 & 0.55 & 2.19 \\
    \hline
    GLM-4.5 & 0.60 & 2.50 \\
    \hline
    Qwen3-32B & 0.28 & 2.79 \\
    \hline
    QwQ-32B & 0.28 & 0.84 \\
    \hline
    qwen3-1.7b & 0.042 & 0.42 \\
    \hline
  \end{tabular}
  \label{tab:model_token_pricing}
\end{table}

\end{document}